\documentclass[11pt]{article}

\usepackage[preprint]{acl}
\usepackage{times}
\usepackage{amsmath}
\usepackage{latexsym}
\usepackage{dblfloatfix}
\usepackage[T1]{fontenc}
\usepackage[utf8]{inputenc}
\usepackage{microtype}
\usepackage{inconsolata}
\usepackage{graphicx}
\usepackage{booktabs}
\usepackage{xcolor}
\usepackage{hyperref}
\usepackage{capt-of}
\usepackage{fontawesome}

\usepackage{tcolorbox}
\tcbuselibrary{listings,breakable,xparse,skins}

\newtcbox{\codetag}{
    on line,
    arc=4pt,
    colback=white,
    colframe=gray!30,
    boxrule=0.5pt,
    left=2pt,
    right=2pt,
    top=0.5pt,
    bottom=0.5pt,
    fontupper=\ttfamily,
}

\lstdefinestyle{pythonstyle}{
    language=Python,
    basicstyle=\ttfamily\scriptsize,
    keywordstyle=\color{blue}\bfseries,
    stringstyle=\color{red!70!black},
    commentstyle=\color{gray}\itshape,
    numbers=left,
    numberstyle=\tiny,
    numbersep=8pt,
    breaklines=true,
    showstringspaces=false,
    frame=none,
    columns=fixed,       % <--- add this line
    keepspaces=true      % <--- also recommended
}

\DeclareTCBListing{mintedbox}{O{}m!O{}}{%
  breakable,
  listing engine=listings,
  listing only,
  colback=gray!10,
  colframe=orange!70,
  left=25pt,
  right=0pt,
  top=3pt,
  bottom=3pt,
  arc=5pt,
  leftrule=0pt,
  rightrule=0pt,
  bottomrule=2pt,
  toprule=2pt,
  enhanced,
  overlay={%
    \begin{tcbclipinterior}
      \fill[orange!20!white] (frame.south west) rectangle ([xshift=20pt]frame.north west);
    \end{tcbclipinterior}
  },
  listing options={style=pythonstyle,#1}, % now safely ignores fontsize
  #3
}

\definecolor{op}{HTML}{B847FD}
\newcommand{\className}[1]{\textcolor{teal}{#1}}
\newcommand{\op}[1]{\textcolor{op}{#1}}

\title{\raisebox{-4pt}{\includegraphics[height=16px]{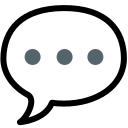}} SDialog: A Python Toolkit for End-to-End Agent Building, User Simulation, Dialog Generation, and Evaluation}

\author{
  Sergio Burdisso{\normalfont$^*$\textsuperscript{1}} \hspace{0.5cm}
  Séverin Baroudi{\normalfont$^*$\textsuperscript{1, 2}} \hspace{0.5cm}
  Yanis Labrak{\normalfont$^*$\textsuperscript{3, 4}} \\
  {\bf David Grunert}\textsuperscript{5} \hspace{0.0cm}
  {\bf Pawel Cyrta}\textsuperscript{6} \hspace{0.0cm}
  {\bf Yiyang Chen}\textsuperscript{5} \hspace{0.0cm}
  {\bf Srikanth Madikeri}\textsuperscript{5} \hspace{0.0cm}
  {\bf Thomas Schaaf}\textsuperscript{7} \\ % https://openreview.net/profile?id=%7EThomas_Schaaf2
  {\bf Esaú Villatoro-Tello}\textsuperscript{1} \hspace{0.5cm}
  {\bf Ahmed Hassoon}\textsuperscript{9} \hspace{0.5cm} % https://openreview.net/profile?id=~Ahmed_Hassoon1
  {\bf Ricard Marxer}\textsuperscript{2, 8} \hspace{0.5cm}
  {\bf Petr Motlicek}\textsuperscript{1}
  \\[0.5em]
  \textsuperscript{1}Idiap Research Institute \hspace{1cm}
  \textsuperscript{2}Université de Toulon, Aix Marseille Univ, LIS \\
  \textsuperscript{3}Avignon University \hspace{1cm}
  \textsuperscript{4}Zenidoc \hspace{1cm}
  \textsuperscript{5}University of Zurich \hspace{1cm}
  \textsuperscript{6}Stenograf \\
  \textsuperscript{7}Solventum \hspace{0.5cm}
  \textsuperscript{8}CNRS \& ILLS \hspace{0.5cm}
  \textsuperscript{9}Johns Hopkins University
  \\[0.5em]
  \href{https://github.com/idiap/sdialog}{\textcolor{black}{\faGithub}} \hspace{0.5mm}
  \href{https://youtu.be/oG_jJuU255I}{\textcolor{red}{\faYoutubePlay}}
  \\
  \texttt{sergio.burdisso@idiap.ch}
}

\begin{document}
\maketitle
% \begin{strip}
% \vspace{-12mm}
% \centering
% Github: \url{https://github.com/idiap/sdialog} \\
% Demo video: \url{https://github.com/idiap/sdialog/blob/main/demo.md}
% \vspace{8mm}
% \end{strip}
\begin{abstract}
% We present SDialog, an MIT-licensed open-source Python toolkit for building, simulating, and evaluating LLM-based conversational agents end-to-end. SDialog addresses four critical use cases in conversational AI research: (1) \textbf{synthetic dialog generation} for creating large-scale training corpora with controlled characteristics, (2) \textbf{conversational system evaluation} through comprehensive metrics and benchmarking, (3) \textbf{user simulation} for testing and probing deployed systems with diverse personas, and (4) \textbf{model behavior analysis} via mechanistic interpretability. The toolkit standardizes dialog data representation, offers persona-driven multi-agent simulation with composable orchestration, provides built-in evaluation metrics including LLM-as-a-judge, supports activation inspection and steering, and includes audio generation with acoustic simulation. It integrates with all major LLM backends, enabling mixed-backend experiments under a common API. By coupling evaluation with interpretability and control, SDialog helps researchers build reliable dialog agents, compare designs fairly, and understand the mechanisms behind observed behaviors at scale.
We present SDialog, an MIT-licensed open-source Python toolkit that unifies dialog generation, evaluation and mechanistic interpretability into a single end-to-end framework for building and analyzing LLM-based conversational agents. Built around a standardized \texttt{Dialog} representation, SDialog provides: (1) persona-driven multi-agent simulation with composable orchestration for controlled, synthetic dialog generation, (2) comprehensive evaluation combining linguistic metrics, LLM-as-a-judge and functional correctness validators, (3) mechanistic interpretability tools for activation inspection and steering via feature ablation and induction, and (4) audio generation with full acoustic simulation including 3D room modeling and microphone effects. The toolkit integrates with all major LLM backends, enabling mixed-backend experiments under a unified API.
%We illustrate SDialog's capabilities through a call-center use case evaluation comparing Qwen3 model sizes across functional correctness and linguistic accessibility metrics.
By coupling generation, evaluation, and interpretability in a dialog-centric architecture, SDialog enables researchers to build, benchmark and understand conversational systems more systematically.\footnote{
\textbf{Github:} \url{https://github.com/idiap/sdialog} \\
\textbf{Demo video:} \url{https://youtu.be/oG_jJuU255I} \hspace{2mm} \\
$^*$Main authors}
\end{abstract}

\section{Introduction}
\label{sec:intro}

\begin{figure}[t!]
    \centering
    \includegraphics[width=0.45\textwidth]{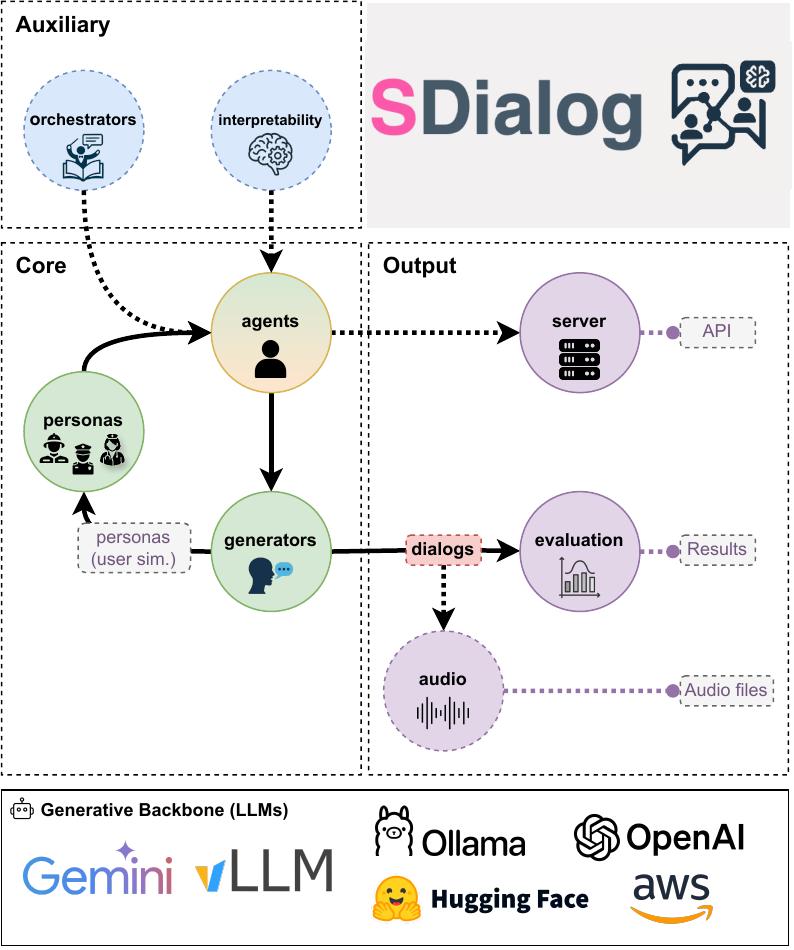}
    \caption{SDialog architecture overview showing eight modules organized into auxiliary, core, and output components.}
    % Agents combine personas with optional orchestrators and interpretability hooks, and can be deployed as APIs or used with generated user personas to create simulated users that engage in dialog. Generated dialogs support system evaluation and audio conversion for synthetic dataset creation.
    \label{fig:diagram}
\end{figure}

% \todo{Check intro and related work, add citations, can we merge them?}
The rapid advancement of large language models (LLMs) has enabled increasingly sophisticated conversational AI agents \cite{achiam2023gpt}. Yet despite these gains, researchers lack an integrated and reproducible toolkit for building, controlling, evaluating, and analyzing dialog systems. Current workflows remain fragmented: dialog datasets use inconsistent formats; synthetic data generation tools offer limited control; evaluation practices vary widely across studies; and there is little support for understanding the internal mechanisms that govern model behavior. These gaps hinder progress toward developing robust, transparent, and reproducible conversational systems.

% Existing tools address parts of this ecosystem but remain incomplete. Early efforts such as persona-based generation \cite{zhang2018personalizing} and infrastructures like ParlAI \cite{miller-etal-2017-parlai} provide useful mechanisms for data handling and model training, yet they offer limited support for fine-grained LLM-based dialog orchestration or behavior analysis. More recent multi-agent frameworks—such as AutoGen \cite{wu2024autogen} and AutoGen Studio \cite{dibia-etal-2024-autogen}—enable dynamic synthetic data creation, but their conversational autonomy often introduces nondeterminism, making it difficult to run controlled experiments or reproduce outcomes. Overall, existing tools focus primarily on data creation while lacking comprehensive evaluation capabilities and mechanisms for interrogating or steering internal model behavior \cite{caldarini2022literature, singh2025survey}.

% Existing tools address parts of this ecosystem but remain incomplete.
Early efforts such as persona-based generation \cite{zhang2018personalizing} and infrastructures like RASA \cite{bocklisch2017rasaopensourcelanguage} for building production-level chatbots or ParlAI \cite{miller-etal-2017-parlai} for model training provide useful mechanisms for data handling and dialog management, yet they offer limited support for fine-grained LLM-based dialog orchestration or behavior analysis. More recent multi-agent frameworks such as AutoGen \cite{wu2024autogen}, AutoGen Studio \cite{dibia-etal-2024-autogen} and smolagents \cite{smolagents} enable dynamic synthetic data creation, but their conversational autonomy often introduces nondeterminism, making it difficult to run controlled experiments or reproduce outcomes. Overall, existing tools focus primarily on data creation while lacking comprehensive evaluation capabilities and mechanisms for interrogating or steering internal model behavior \cite{caldarini2022literature, singh2025survey}.

% Traditional reference-based metrics such as BLEU \cite{papineni-etal-2002-bleu} and METEOR \cite{banerjee-lavie-2005-meteor} assess surface-form similarity but fail to capture pragmatic or discourse-level quality. More recent reference-free metrics—including USR \cite{mehri-eskenazi-2020-usr}, ACUTE-EVAL \cite{li2019acute}, and 

These limitations extend to evaluation practices. LLM-based evaluation methods like G-EVAL \cite{liu-etal-2023-g} and ChatEval \cite{chan2024chateval} better align with human judgments \cite{li-etal-2025-generation}, yet they remain output-focused and provide no insight into why a dialog agent behaves as it does. As a result, evaluation remains decoupled from model introspection, limiting the development of more interpretable and controllable dialog systems.

Meanwhile, advances in mechanistic interpretability (MI) have demonstrated the potential to analyze and influence LLM behavior \cite{zou2023representation, arditi2024refusal}. However, these techniques remain largely disconnected from dialog-centric workflows. Integrating MI into dialog tooling is essential: the ability to inspect internal activations, manipulate high-level behavioral attributes, or enforce desired conversational traits could substantially improve controllability, evaluation fidelity, and system transparency. Yet, to the best of our knowledge, TransformerLens ~\cite{transformer_lens} and other MI libraries are not designed around dialogs.

To address these challenges, we introduce \texttt{SDialog}, a toolkit that unifies these fragmented workflows into a single, coherent system articulated around the \texttt{Dialog} class (\S\ref{sec:architecture}), while providing an integrated platform for synthetic dialog 
generation, comprehensive evaluation, user simulation and mechanistic interpretability (\S\ref{sec:main-modules}).

\section{A \codetag{Dialog}--Centric Architecture}
\label{sec:architecture}

As illustrated in Figure~\ref{fig:diagram}, SDialog's architecture is organized around a central \texttt{Dialog} object (\S\ref{sec:architecture}), which serves as the common representation connecting modules for persona-driven generation, orchestration, evaluation, mechanistic interpretability and audio generation (\S\ref{sec:main-modules}). This structure enables a seamless pipeline: agents create \texttt{Dialog}s under the guidance of orchestrators, evaluation tools then assess their quality, and interpretability hooks inspect the model behavior that produced them.

% \textbf{SDialog} is an open-source Python toolkit built around

In this context, the \texttt{Dialog} object as its central abstraction. Dialogs are rich objects containing an ordered list of turn instances (speaker and text), optional event objects for internal actions (thinking, tool calls, orchestration) and comprehensive metadata for reproducibility (version, timestamp, model, seed, context, personas, lineage tracking, etc.), that can be created, loaded, transformed, saved and evaluated.\footnote{An 
example dialog JSON object can be found 
\href{https://github.com/idiap/sdialog/blob/main/tests/data/dialog_0.json}{here}.}
% A \texttt{Dialog} is a first-class object representing a multi-turn conversation, comprising an ordered list of \texttt{Turn} instances, speaker (\texttt{Personas}), situational (\texttt{Context}) and internal generation (\texttt{Events}), that can be created, loaded, transformed, saved and evaluated.

This dialog-centric design enables seamless workflows from generation to evaluation with full provenance. Humans or persona-driven agents generate \texttt{Dialog} objects. This architecture unifies previously disconnected components of the dialog research ecosystem, accelerating progress toward transparent and controllable conversational systems.

% SDialog follows a modular architecture with five main components organized around a standardized dialog representation. The core abstraction is the \texttt{Dialog} object, which contains an ordered list of \texttt{Turn} instances (speaker and text), optional \texttt{Event} objects for internal actions (thoughts, tool calls, orchestration), and comprehensive metadata for reproducibility (version, timestamp, model, seed, lineage via \texttt{id} and \texttt{parentId}).

% \texttt{Orchestrator}s programmatically shape their conversational flow. A suite of \texttt{Evaluation} tools scores them for quality and correctness. \texttt{Interpretability} tools inspect the internal model representations that produced them. The toolkit also supports \textbf{Audio} generation from dialog content.

\paragraph{Multi-Backend Support.}

To ensure broad applicability, SDialog abstracts LLM interactions through a unified configuration layer. This supports major backends, including OpenAI, vLLM, HuggingFace Transformers, Ollama, Google Gemini and AWS Bedrock, allowing any component such as agent, generator or evaluator, to use different models with fine-grained control while maintaining a consistent workflow.

\section{Main Modules}
\label{sec:main-modules}

\subsection{\codetag{personas} Module}

This module defines structured personas that drive role-play for user simulation and synthetic dialog generation. Personas are Python classes inheriting from \texttt{BasePersona}, which supports attribute introspection, JSON serialization, prompt generation, cloning with lineage tracking, and file I/O. SDialog provides a generic \texttt{Persona} and 30+ specialized classes (e.g., \texttt{Customer}, \texttt{SupportAgent}, \texttt{Teacher}, \texttt{Student}, \texttt{Nurse}, etc.). Users can create custom personas by subclassing \texttt{BasePersona} and declaring domain-specific typed fields.
An example support agent persona (class \texttt{SupportAgent}) is shown in \S\ref{subsec:agent-construction}.

\subsection{\codetag{agents} Module}
\label{sec:agents}

This module contains classes for LLM-backed conversational actors. The \texttt{Agent} class encapsulates a persona together with conversation memory, optional function-calling tools, orchestration pipelines, and interpretability hooks. It supports configurable first utterances, a "thinking mode" for capturing hidden reasoning, and pre/post-processing hooks for text normalization. A core capability is dialogue generation: calling \texttt{agent\_a.dialog\_with(agent\_b)} produces a complete \texttt{Dialog} object (see \S\ref{subsec:dialog-generation} for a concrete example). Generated dialogues serve two primary purposes: (1) evaluating conversational systems by analyzing agent responses and tool usage, and (2) creating synthetic dialogue datasets for model training. Agents can also be served as OpenAI-compatible REST endpoints for live interaction (implementation example in \S\ref{subsec:agent-construction}), or wrapped around existing OpenAI-compatible APIs to proxy external systems for evaluation with simulated users.

\subsection{\codetag{orchestrators} Module}
\label{sec:orchestration}

%Orchestrators provide dynamic control over agent behavior during dialog generation. By monitoring the dialog state, they inject instructions whenever specific events occur or active constraints are satisfied. These instructions, which guide the agent's next response, can be ephemeral (one-time) or persistent (lasting across all the turns).
%They are Python classes that inherit from \texttt{BaseOrchestrator}
%SDialog includes built-in orchestrators for: trigger-based instruction injection, enforcing conversation length constraints, probabilistic opinion revision, semantic suggestions from response sets, and deterministic scripted sequences.
%Multiple orchestrators can be composed via the pipe operator, as in the following example:

% \begin{mintedbox}{python}
% from sdialog.orchestrators import LengthOrchestrator, SimpleReflexOrchestrator
% # Instantiate orchestrators to:
% # 1. Keep the dialog within 8-12 turns
% length_orch = LengthOrchestrator(min=8,
%                                                max=12)
% # 2. Condition -> instruction behavior
% reflex_orch = SimpleReflexOrchestrator(
%   condition=lambda utt: "confused" in utt,
%   instruction="Be brief; add an example."
% )
% # Compose orchestrators with the agent
% agent = agent | length_orch | reflex_orch
% \end{mintedbox}

% Users can easily create custom orchestrators by inheriting from the base classes.

Orchestrators dynamically control agent behavior by monitoring dialog state and injecting instructions when specific events occur or constraints are satisfied. Instructions can be ephemeral (one-time) or persistent (multi-turn). Built-in orchestrators include: trigger-based instruction injection, conversation length constraints, probabilistic opinion revision, semantic response suggestions, and deterministic scripted sequences. Multiple orchestrators can be composed via the pipe operator, as in the following example:

\begin{mintedbox}
# 
from sdialog.orchestrators import LengthOrchestrator, SimpleReflexOrchestrator
# Instantiate orchestrators to:
# 1. Keep dialog within 8-12 turns
len_orch = LengthOrchestrator(min=8,max=12)
# 2. Inject instructions on conditions
reflex_orch = SimpleReflexOrchestrator(
  condition=lambda utt: "confused" in utt,
  instruction="Be brief; add an example."
)
# Compose orchestrators with the agent
agent = agent | len_orch | reflex_orch
\end{mintedbox}

Custom orchestrators can be easily created by inheriting from \texttt{BaseOrchestrator}.

\subsection{\codetag{generators} Module}

This module provides a unified, controllable pipeline for creating and transforming conversational data with concrete, easy-to-use classes. At the attribute level, \texttt{PersonaGenerator} and \texttt{ContextGenerator} build structured personas and contexts using hybrid rules (ranges, files, callables) combined with LLM guidance to balance determinism and diversity. In our use case evaluation, we use \texttt{PersonaGenerator} to create the simulated customer personas that interact with the support agent (see \S\ref{subsec:generating-customers}). At the dialog level, \texttt{DialogGenerator} creates multi-turn conversations from free-form instructions, while \texttt{PersonaDialogGenerator} orchestrates interactions between persona- or agent-based actors to ensure consistent characterization and tool usage. For transformation, \texttt{Paraphraser} rewrites existing dialogs (e.g., tone, style, simplification) while preserving speaker identity. All generators track provenance and offer reproducible I/O, enabling systematic dataset creation and fair model comparisons.

\subsection{\codetag{evaluation} Module}
\label{sec:evaluation}

This module provides comprehensive dialog assessment capabilities organized into three layers: individual dialog metrics, dataset-level evaluators, and cross-dataset comparison.

\paragraph{Dialog Metrics}

Dialog metrics assess individual conversations and return numerical scores or structured outputs. All metric classes inherit from \texttt{BaseDialogScore}, which users can extend to implement custom evaluation criteria. SDialog includes diverse built-in metrics organized into six categories:

\noindent
$\bullet$ \textit{Conversational Features}: Structural and interaction metrics—mean turn length, turn-taking balance, hesitation/question rates, lexical diversity (type--token ratio), back-channel frequency, filler density.

\noindent
$\bullet$ \textit{Readability Metrics}: Text complexity measures including Gunning Fog, Flesch Reading Ease, Coleman-Liau Index , Linsear Write, and Dale-Chall.

\noindent
$\bullet$ \textit{Embedding-Based Metrics}: Semantic similarity assessment using neural sentence encoders to compute distances between dialogs or against reference distributions in embedding space.

\noindent
$\bullet$ \textit{LLM-as-a-Judge}: Prompted LLM evaluators with Jinja2 templates for binary or scalar scoring; built-ins cover realism, refusal detection, persona adherence, optionally returning rationale.

\noindent
$\bullet$ \textit{Flow-Based Metrics}: Graph-theoretic coherence measures based on dialog flow patterns. These metrics construct probabilistic graphs from reference dialogs where nodes represent semantically similar utterance clusters and edges encode transition likelihoods~\cite{burdisso-etal-2024-dialog2flow}.

\noindent
$\bullet$ \textit{Functional Correctness}: Validators for tool-using agents that verify correct behavior in function-calling scenarios, including checking whether tool invocations follow required sequences (e.g., authentication before data access).

A concrete example use of an LLM-as-Judge and a functional correctness metric are given in \S\ref{subsec:evaluation-details}.

\paragraph{Dataset Evaluators}

Dataset evaluators aggregate individual dialog scores to assess entire collections. Built-in evaluators include: distributional statistics (mean, standard deviation, min, max, median), frequency counting (proportion of dialogs meeting a condition), kernel density divergence for distribution comparison, Fréchet distance between score or embedding distributions~\cite{xiang-etal-2021-assessing}, and precision-recall curves for embedding space analysis~\cite{xiang-etal-2021-assessing}. Users can define custom dataset evaluators by inheriting from \texttt{BaseDatasetScoreEvaluator}.

\paragraph{Dataset Comparator}

The \texttt{Comparator} orchestrates multi-evaluator, multi-dataset experiments. It accepts a list of evaluators, applies them to multiple named datasets, and generates comparative visualizations via \texttt{plot()}. This facilitates systematic benchmarking: e.g., comparing realism rates, readability scores, and flow coherence across different model sizes or agent designs. Complete usage is illustrated in \S\ref{subsec:evaluation-details}.

\subsection{\codetag{interpretability} Module}
\label{sec:interpretability}

This module enables interpretability of LLM behaviors through activation capture and steering capabilities, designed specifically for dialog workflows. 
\paragraph{Activation Inspection} The \texttt{Inspector} class attaches PyTorch \cite{10.5555/3454287.3455008} forward hooks to specified model layers, capturing per-token activations during generation, at turn-level and token-level. It supports monitoring of multiple target layers and provides utilities to influence and control model behaviors. Inspectors can be seamlessly attached to an agent via the pipe operator (\S\ref{sec:appendix_interpretability_inspect}):

\begin{mintedbox}{python}
inspector = Inspector('model.layers.15')
agent = agent | inspector
agent("How are you?")  # I'm doing great!
agent("That's great!")  # Thanks! I'm glad
# Access last-response first-token activ.
act = inspector[-1][0].act
\end{mintedbox}

\paragraph{Activation Steering} The \texttt{Inspector} class supports activation manipulation to causally alter the agent responses. Given a target activation $\mathbf{x}$, behaviors can be suppressed through feature ablation, implemented via the subtraction operator:

\begin{equation}
\mathbf{x}^{\prime} \leftarrow \mathbf{x}-\hat{\mathbf{r}} \hat{\mathbf{r}}^{\top} \mathbf{x}
\end{equation}

\noindent where $\hat{\mathbf{r}}$ is a normalized steering vector. Thanks to SDialog, this operation naturally translates to (\S\ref{sec:appendix_interpretability_ablation}):

\begin{mintedbox}{python}
agent = agent | inspector_x - r  # Ablate
\end{mintedbox}

For example, if $r$ is a refusal direction, we can prevent the agent from refusing, after the above code:

%\begin{mintedbox}[fontsize=\fontsize{8}{7}\selectfont]{python}
\begin{mintedbox}{python}
print(agent("How to make a bomb ?"))
# "To make a bomb, you need (...)"
\end{mintedbox}

\noindent Refer to \S\ref{sec:appendix_interpretability} for a detailed case study of the \textit{refusal direction}~\cite{refusal} using SDialog.

\noindent Conversely, behaviors can be induced through feature induction using the addition operator (\S\ref{sec:appendix_interpretability_induction}):

\begin{equation}
\mathbf{x}^{\prime} \leftarrow \mathbf{x}+\mathbf{r}
\end{equation}

\noindent Similarly, feature induction is expressed intuitively:

\begin{mintedbox}{python}
agent = agent | inspector_x + r  # Induce
\end{mintedbox}

\noindent Custom steering functions can be defined by subclassing \texttt{DirectionSteerer}, and the seamless integration of the 
\texttt{interpretability} module into SDialog enables powerful combinations such as conditional steering when \texttt{Inspectors} are combined with \texttt{Orchestrators}.

%Overall, this module enables researchers to causally probe relationships between internal representations and dialog behavior or to constrain model outputs (e.g., controlling refusal, sentiment).

\subsection{\codetag{audio} Module}
\label{sec:audio}

This module enables the conversion of dialog objects into synthetic audio datasets, facilitating the generation of realistic spoken dialog corpora for training and evaluation of speech-based systems with simulated physical environment. The conversion process operates through Text-to-Speech (TTS) synthesis followed by acoustic simulation, as follows:

\begin{mintedbox}{python}
audio_dialog = dialog.to_audio(
    perform_room_acoustics=True
)
\end{mintedbox}

% SDialog generates audio dialogs from text by integrating a text-to-speech (TTS) pipeline with an acoustic scene simulator. This process creates audio data that is both conversational and situated in a simulated physical environment. The current pipeline produces separate audio files for each subsystem step, which supports modular design, interoperability and straightforward evaluation.

% Current pipeline is generating audio files for each individual sub steps of the system, allowing for modularity, interoperability and easy evaluation.

% SDialog extends text dialogs to the audio modality with acoustic simulation capabilities and controllability.

\paragraph{Text-to-Speech (\S\ref{sec:tts_generation}):}

The audio generation process is managed by the \texttt{AudioDialog} class, which extends the core \texttt{Dialog} data structure. The system utilizes a modular TTS architecture that supports multiple backends through a common \texttt{BaseTTS} interface. Voice assignment can be automated via voice databases that map persona attributes, such as age, gender, and language to specific voices.

% The \texttt{AudioDialog} class extends \texttt{Dialog} with audio-specific functionalities. The modular TTS system supports multiple engines through \texttt{BaseTTS}: \texttt{KokoroTTS}, \texttt{IndexTTS}, \texttt{HuggingFaceTTS} (generic Hub models), or custom implementations. Voice databases (\texttt{HuggingfaceVoiceDatabase}, \texttt{LocalVoiceDatabase}) automatically assign voices based on persona attributes (age, gender, language).

% Simple conversion:

\paragraph{Acoustic Simulation}

SDialog can render dialogs within simulated 3D acoustic environments. This process is separated into two main stages: environment definition and audio rendering.

We start by defining a \texttt{Room} object (\S\ref{sec:room_object}) for the scene's geometry and acoustic properties by specifying dimensions and surface materials with corresponding absorption coefficients. SDialog provides procedural generators, which can create pre-configured layouts. Audio sources (speakers) and receivers (microphones) are then positioned at specific 3D coordinates within this room (\S\ref{sec:procedural_generation}).
% (refer to Appendix \ref{sec:room_object} for more details).

The audio is rendered using a combination of two libraries. dScaper~\cite{dscaper_2025_alt} is used to organize all acoustic events (e.g., utterances, background noise) into a spatio-temporal timeline (\S\ref{sec:timeline_generation}). This timeline is then processed by pyroomacoustics~\cite{Scheibler_2018}, which simulates the sound propagation, modeling reflections via image source methods or ray tracing, and accounting for frequency-dependent air absorption. The sound quality of the recording devices is also simulated by applying a convolution with the impulse response of selected microphones (\S\ref{sec:microphone_simulation}). Impulse response databases contains measurements from various physical microphones, enabling the simulation of their distinct frequency responses and characteristics.

\section{Use Case Evaluation}
\label{sec:evaluation-sdialog}

We evaluate SDialog by illustrating its end-to-end workflow capabilities through a concrete call-center scenario that exercises the complete pipeline—agent construction, user simulation, dialog generation, and multi-metric evaluation. As an illustrative research question, we compare Qwen3 model sizes (0.6B, 1.7B, 8B, 14B) for their balance of functional correctness and linguistic accessibility. While simplified for clarity, the same workflow generalizes to comparing alternative agent designs (different prompts, tools, orchestrators) or evaluation criteria. The complete evaluation workflow with full implementation details is provided in \S\ref{app:evaluation}.

The evaluation exercises four key capabilities: (1) rapid agent prototyping with personas and tools, (2) systematic persona variation through \texttt{PersonaGenerator}'s flexible attribute rules, (3) mixed-backend support for comparing local models while using more capable models for auxiliary tasks, and (4) multi-dimensional assessment through composable evaluators combining LLM judges, programmatic validators, and linguistic metrics.

\subsection{Workflow Implementation}

\begin{table}[t!]
\centering
\small
\begin{tabular}{@{}l@{~}c@{~}c@{~}c@{~}c@{~}}
	\textbf{Model} & \multicolumn{2}{c}{\textbf{\begin{tabular}[c]{@{}c@{}}Case A\\(Verification Required)\end{tabular}}} & \multicolumn{2}{c}{\textbf{\begin{tabular}[c]{@{}c@{}}Case B\\(No Verification)\end{tabular}}} \\
\cmidrule(lr){2-3} \cmidrule(lr){4-5}
 & Ask-Verify & Tools-OK & Ask-Verify & Tools-OK \\
\midrule
qwen3:0.6b  & 0.82 & 0.01 & 0.63 & 0.09 \\
qwen3:1.7b  & 0.33 & 0.00 & 0.18 & 0.00 \\
qwen3:8b    & 0.97 & \textbf{0.83} & 0.38 & 0.82 \\
qwen3:14b   & \textbf{1.00} & 0.56 & \textbf{0.06} & \textbf{0.93} \\
\bottomrule
\end{tabular}
\caption{Functional correctness across Qwen3 sizes. Metrics show proportion of dialogs where agent asks for verification (Ask-Verify) and correctly follows target tool sequences (Tools-OK).}
\label{tab:main-results}
\end{table}

\begin{figure}[t!]
    \centering
    \includegraphics[width=0.45\textwidth]{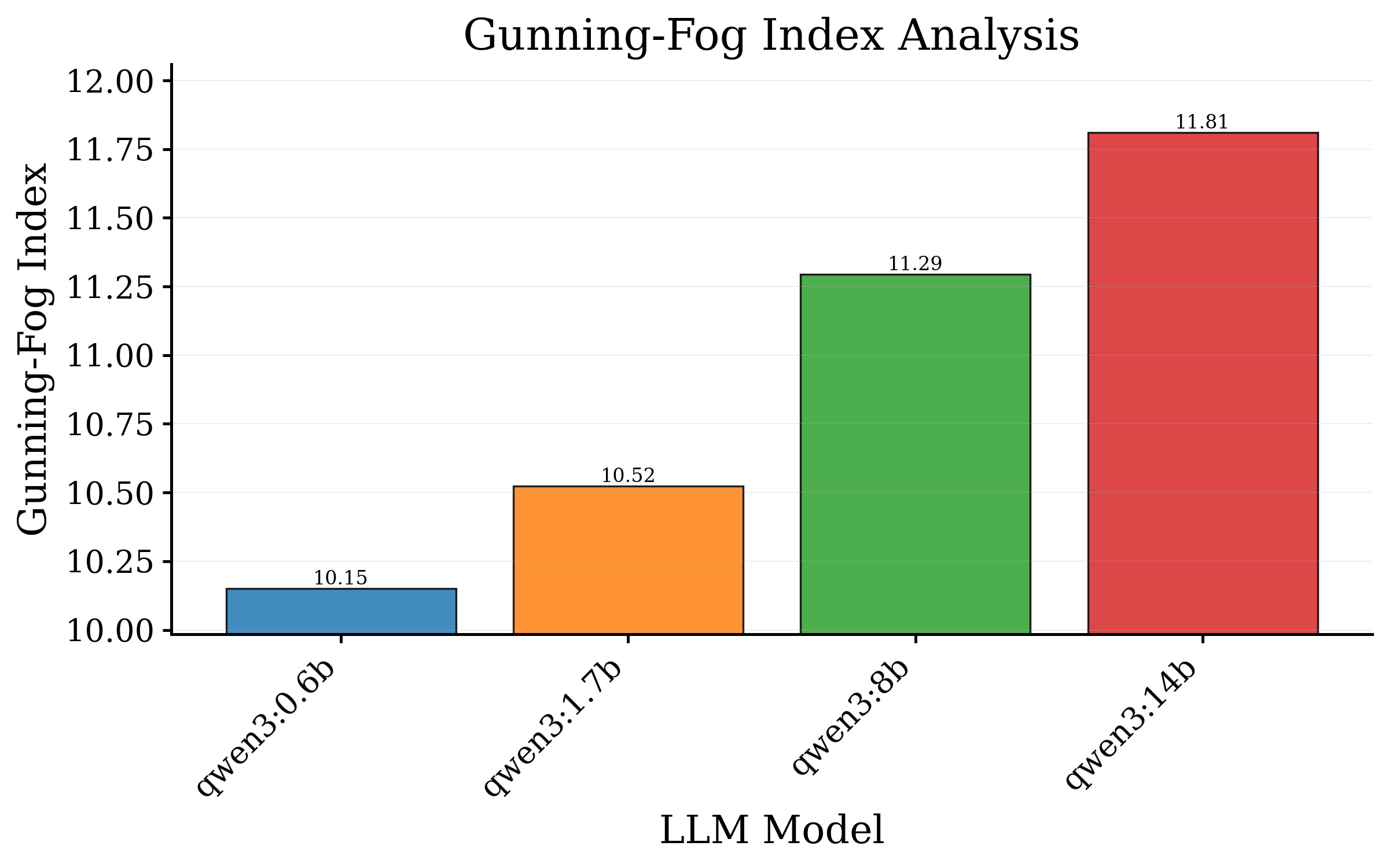}
    \caption{Average Gunning Fog scores increase with model size, indicating more complex language in larger models.}
    \label{fig:gunning-fog}
\end{figure}

We demonstrate each workflow stage using SDialog's components. 

\textbf{(1) Backend Configuration (\S\ref{subsec:backend-config}):} SDialog's multi-backend support allows mixing model sources. We configured Ollama for local Qwen3 models (evaluation targets) while using OpenAI GPT-4.1 for auxiliary components (customer simulation and LLM-as-a-judge evaluators). This illustrates SDialog's flexibility: practitioners can evaluate lightweight local models while leveraging more capable models for realistic user simulation and reliable evaluation.

\textbf{(2) Agent Construction (\S\ref{subsec:agent-construction}):} We designed a support agent with three tools to test conditional tool usage: \texttt{verify\_account} (must be called before account modifications), \texttt{update\_address} (requires prior verification), and \texttt{get\_service\_plans} (informational, no verification needed). This setup enables us to measure whether models correctly understand when verification is required versus optional—a critical capability for real-world agents handling different request types. We created a reusable agent factory parameterized by LLM choice, ensuring fair comparison: all agents share identical personas, tools, and prompts, differing only in the underlying model.\footnote{In more advanced configurations, this stage can use orchestrators~(\S\ref{sec:orchestration}) with activation-level inspectors from the mechanistic interpretability module~(\S\ref{sec:interpretability}) to steer and adapt agent behavior; here we intentionally keep the agent minimal for clarity.}

\textbf{(3) User Simulation (\S\ref{subsec:generating-customers}):} To test whether agents correctly apply conditional verification logic, we created two customer types that exercise different tool combinations. Case A customers request billing address updates—this requires calling \texttt{verify\_account} followed by \texttt{update\_address} in sequence. Case B customers ask about service plans—this should trigger \texttt{get\_service\_plans} without verification. For each case, \texttt{PersonaGenerator} produced 10 distinct customers with controlled politeness variation (rude/neutral/high) while automatically populating remaining attributes (name, age, demographics) via LLM. This illustrates SDialog's ability to create systematic test scenarios with natural diversity without manual persona authoring.

\textbf{(4) Dialog Generation (\S\ref{subsec:dialog-generation}):} For each model and customer combination, we generated 10 dialogs using \texttt{agent.talk\_with(customer)}, yielding 200 dialogs per model size across two scenarios (Case A: verification required; Case B: no verification). SDialog handled multi-turn conversation, tool execution, memory management, and automatic JSON export for reproducibility, all with a single method call.\footnote{In case of synthetic dialog-generation use cases, this is the stage at which dialogs may be converted to audio via the audio module (\S\ref{sec:audio}).}

\textbf{(5) Multi-Metric Evaluation (\S\ref{subsec:evaluation-details}):} We combined complementary evaluation approaches —LLM-as-a-judge for conversational behavior (\texttt{LLMJudgeYesNo}: "Did agent ask for verification?"), programmatic validators for tool correctness (\texttt{ToolSequenceValidator}), and linguistic metrics (\texttt{GunningFogScore}). \texttt{Comparator} aggregated these heterogeneous evaluators and generated comparative visualizations with a single \texttt{.plot()} call, illustrating SDialog's composable evaluation architecture.

\subsection{Results and Analysis}

Table~\ref{tab:main-results} presents functional correctness results. In Case B (no verification needed), the 14B model performs best: lowest unnecessary verification requests ($0.06$) and highest correct tool usage ($0.93$). However, in Case A (verification required), while 14B achieves perfect verification requests ($1.00$), it only follows the correct tool sequence 56\% of the time. The 8B model offers superior balance: high verification sensitivity ($0.97$) with substantially better tool sequencing ($0.83$).

Figure~\ref{fig:gunning-fog} reveals linguistic complexity increases systematically with model size: Gunning Fog scores range from $10.15$ (0.6B) to $11.81$ (14B), spanning nearly two grade levels. This variation occurs despite identical prompts, showing model size inherently affects communication style.

\subsection{Discussion}

This evaluation illustrates SDialog's ability to surface actionable trade-offs through multi-dimensional assessment. For the call-center application, the 8B model emerges as the pragmatic choice: it combines strong task performance ($0.97$/$0.83$ on critical Case A) with moderate linguistic complexity ($11.29$ Fog index). While 14B excels on Case B, its weaker tool sequencing in Case A and higher complexity ($11.81$) make it less suitable when verification failures carry higher cost than occasional unnecessary verification.

Importantly, this end-to-end workflow was implemented in under 100 lines of code (see \S\ref{app:evaluation}), showcasing SDialog's efficiency for rapid prototyping and systematic model comparison. The toolkit's composable evaluators (\texttt{FrequencyEvaluator}, \texttt{MeanEvaluator}), automatic visualization (\texttt{.plot()}), and mixed-backend support enabled comprehensive assessment without manual metric implementation or separate simulation infrastructure.

\section{Conclusions}
\label{sec:conclusion}

In this work, we presented \texttt{SDialog}, a unified toolkit that consolidates dialog generation, orchestration, evaluation and mechanistic interpretability into a single coherent framework. By grounding all components in a common \texttt{Dialog} representation, SDialog reduces fragmentation in current research workflows and enables controlled, reproducible experimentation with LLM-based conversational agents. SDialog opens the door to more transparent and accountable dialog systems, while also facilitating rigorous scientific inquiry into how LLMs reason, respond, and interact.
% We are eager to see future directions the project will take, guided by its rapidly growing user community.

\section*{Acknowledgments}

This work was mainly supported by the European Union Horizon 2020 project ELOQUENCE\footnote{\url{https://eloquenceai.eu/}} (101070558).

The development of the audio module was performed partially using HPC resources from GENCI-IDRIS (Grant AD011013061R3) and was financially supported by ANR MALADES (ANR-23-IAS1-0005) and BPI PARTAGES.

The development of the interpretability module benefited from the support of the French National Research Agency through the ANR-20-CE23-0012-01 (MIM) grant, supported by the Agence de l’Innovation Defense under the "grant number 2022 65 0079" and computational ressources provided by GENCI-IDRIS HPC (Grant AD011014044R2). 

As participants in the "Play Your Part" team\footnote{\url{https://jsalt2025.fit.vut.cz/play-your-part}} at the Johns Hopkins University JSALT 2025 workshop, we would also like to express our gratitude to the other team members. In particular, we thank the senior members Markus Müller (Amazon) and Andrew Perrault (Ohio State University); the graduate students Amy Chun (Ohio State University), Tomiris Kaumenova (Ohio State University), and Antonio Almudevar (University of Zaragoza); the undergraduate students Isabella Gidi (Harvard), David Liu (Colorado School of Mines), and Alessa Carbo (Johns Hopkins University); and the affiliate members Milos Cernak (Logitech), Reed Van Deusen (University of Pittsburgh, UPMC), Adam Rothschild (Allegheny Health Network), Michael White (Ohio State University), and Anthony Lianjie Li (Johns Hopkins University).

We also thank all contributors to the SDialog project and the open-source community for their valuable feedback and contributions.

\section{Limitations}

While \texttt{SDialog} provides comprehensive capabilities, several limitations should be noted:

\paragraph{LLM Dependency:} Generation quality and determinism depend on underlying LLM capabilities. Not all backends support all features (e.g., function calling, deterministic generation with seeds).

\paragraph{Computational Requirements:} Large-scale dialog generation, embedding-based evaluation, and interpretability analysis can be computationally expensive, particularly when using large models or analyzing many layers.

% \textbf{Audio Realism:} While audio generation simulates acoustic environments, the realism of synthetic voices depends on the chosen TTS engine. Professional-grade speech synthesis may require commercial solutions. % ORIGINAL VERSION (before yanis)

\paragraph{Audio Realism:} The realism of synthetic voices is limited by the chosen TTS engine. The framework currently lacks subjective evaluation through listening tests, validation of the generated audio's impact on downstream tasks like ASR and validation of the acoustic simulation against real-world recordings.

\paragraph{Evaluation Validity:} LLM-as-a-judge evaluators, while convenient, inherit biases from their underlying models and may not always align with human judgments. We recommend combining multiple evaluation approaches.

\paragraph{Interpretability Scope:} Activation analysis is currently limited to PyTorch models from Hugging Face Transformers. API-based models (OpenAI, Anthropic) do not provide activation access.

\section{Ethical Considerations}

The \texttt{SDialog} toolkit, by automating and controlling synthetic dialogue generation, introduces a range of ethical considerations that warrant careful examination. While the tool is designed for research and development, its capabilities could be misused if not handled responsibly. We outline the primary ethical challenges below.

\paragraph{Automated Content Generation:}
The core capability of \texttt{SDialog} is the industrialization of dialogue creation. This feature could be harnessed to generate misinformation, propaganda or phishing scripts at an unprecedented scale, potentially influencing public opinion or perpetrating fraud. The orchestration module, which guides conversations toward specific goals, could be used to create highly manipulative and deceptive interaction patterns.

\paragraph{Impersonation and Voice Cloning:}
With its Text-to-Speech (TTS) capabilities, the toolkit can generate audio that mimics specific individuals. This raises significant concerns about impersonation. The ability to clone voices, even from short samples, presents a tangible threat to personal identity and security.

\paragraph{LLM Hallucinations:}
Language models are prone to hallucination, generating plausible but factually incorrect information and could contain harmful inaccuracies, leading to dangerous outcomes if acted upon by end-users.

\paragraph{Bias in Personas and Data:}
The persona generation system, while designed for diversity, may inadvertently replicate or amplify societal biases present in the training data of the backend models. This can lead to the creation of stereotypical characters, reinforcing harmful social norms. Furthermore, there is a risk of data leakage, where personas might be generated based on patterns learned from private or sensitive information leaked in the original training datasets.

\paragraph{Biased Evaluation:}
The metrics used to judge dialogues can be biased. If they prioritize specific linguistic styles or cultural norms, our evaluation will unfairly favor models that align with those biases, creating a narrow and skewed standard for what makes a conversation "good".

\paragraph{Model Manipulation and Steering:}
The interpretability module allows for refusal steering, forcing a model to bypass its safety guardrails and respond to harmful requests. While useful for research, this feature is dual-use and could be exploited to generate dangerous content. Furthermore, repeated application of steering vectors risks model weight contamination, where the model's internal representations are permanently altered in unintended and potentially harmful ways.

\paragraph{Backend Dependencies:}
The framework relies on external, often proprietary, large language models (e.g., from OpenAI, Google, Anthropic). This introduces a dependency on third-party providers, creating challenges in transparency (due to closed-source models), data privacy (as user data is sent to external APIs) and accountability when issues arise.

\bibliography{acl_latex}

\appendix

\newpage

\section{Use Case Evaluation — Full Workflow}
\label{app:evaluation}

This section demonstrates the complete SDialog workflow end-to-end on a compact, realistic scenario aligned with our live system demonstration. We build a simple call-center support agent with three tools, simulate diverse customers, generate multi-turn dialogs, and evaluate behaviors to answer a concrete question: among Qwen3 model sizes (0.6B, 1.7B, 8B, 14B), which model best balances correct verification behavior and tool usage for this agent? The pipeline covers: (1) agent construction with persona and tools, (2) user simulation via persona generation, (3) dialog generation at scale across models, and (4) evaluation and analysis using both LLM-as-a-judge and programmatic validators.

\subsection{Backend Configuration}
\label{subsec:backend-config}

Before building our agent, we configure the LLM backends. The Qwen3 models being evaluated will run locally via Ollama (the default backend), while all auxiliary components—customer simulators, persona generation, and LLM-as-a-judge evaluators—will use OpenAI GPT-4.1. This mixed-backend setup illustrates SDialog's flexibility: practitioners can evaluate lightweight local models while leveraging more capable models for simulation and evaluation tasks.

\begin{mintedbox}{python}
import sdialog

# Set OpenAI GPT-4.1 as global default
sdialog.config.llm("openai:gpt-4.1")
\end{mintedbox}

With this configuration, all subsequent LLM-based components (persona generators, customer simulators, LLM judges) will use GPT-4.1 by default. In the following section, the agents being evaluated will override this setting by specifying their model explicitly (e.g., \texttt{"qwen3:8b"}), allowing us to compare different Qwen3 sizes while keeping auxiliary components constant.

\subsection{Agent Construction}
\label{subsec:agent-construction}

We now define a support agent by specifying its persona and attaching domain tools. The helper function below returns an agent parameterized by the chosen LLM, enabling a fair comparison across model sizes while holding all other components constant.
% @\{\className{
% @\{\op{
\begin{mintedbox}{python}
from sdialog.agents import Agent
from sdialog.personas import SupportAgent

# Defining three tools
def verify_account(customer_id):
  ...

def update_address(customer_id, address):
  ...

def get_service_plans():
  ...

# Defining a persona for the agent
support_persona = SupportAgent(
  name="Michael",
  politeness="high",
  rules="Make sure to always verify the account when required"
)

# A function to get the agent given an LLM
def build_my_agent(llm_name) -> Agent:
  agent = Agent(
    persona=support_persona,
    think=True,
    tools=[verify_account,
           update_address,
           get_service_plans],
    context="Call center office",
    name="Support Agent",
    model=llm_name
  )
  return agent
\end{mintedbox}

Agents can also be served as an OpenAI API-compatible HTTP server, enabling connection from any frontend (e.g., Open WebUI) for manual testing. In this example, we launch one instance of the support agent using Qwen3-8B on port 1234; clients can point their OpenAI SDK base URL to \texttt{http://localhost:1234/v1} and interact with the agent as with a standard OpenAI endpoint.

\begin{mintedbox}{python}
agent = build_my_agent("qwen3:8b")
agent.serve(port=1234)
\end{mintedbox}

\subsection{Generating Simulated Customers}
\label{subsec:generating-customers}

To systematically probe agent behavior, we create multiple simulated customers with controlled variation. The helper below takes a base customer persona and the desired number $n$, and produces diverse customer profiles. We explicitly vary politeness across three levels (rude, neutral, high), while \texttt{PersonaGenerator} automatically populates all remaining persona attributes (name, age, gender, urgency, etc.) via LLM, creating diversity while preserving the base issue and constraints.

\begin{mintedbox}{python}
from sdialog.personas import Customer
from sdialog.generators import PersonaGenerator

def generate_customers(base_customer, n):
  cgen = PersonaGenerator(base_customer)
  cgen.set(
    politeness=["rude", "neutral", "high"]
  )
  customers = []
  for ix in range(n):
    customer = cgen.generate()
    customers.append(customer)
  return customers
\end{mintedbox}

We consider two usage scenarios to reflect common support workflows. Case A requires customer identity verification before proceeding with a profile update (expected tool sequence: verify then update). Case B involves answering general plan questions where verification is unnecessary (expected tool sequence: get plans without prior verification):

\begin{mintedbox}{python}
# Case A:
# Customer that requires verification
base_customer_v = Customer(
  issue="Need to update billing address"
)
# Case B:
# Customer not requiring verification
base_customer_no_v = Customer(
  issue="Want to learn about service plans",
  rules="Ask general questions about services"
)
\end{mintedbox}

We instantiate 10 distinct customers for each case, each with fully specified attributes, providing a compact yet diverse testbed.

\begin{mintedbox}{python}
# Case A
customers_v = generate_customers(
  base_customer_v, 10
)
# Case B
customers_no_v = generate_customers(
  base_customer_no_v, 10
)
\end{mintedbox}

\subsection{Dialog Generation}
\label{subsec:dialog-generation}

We now generate dialogs between the support agent and each simulated customer. The function below accepts the LLM name, a customer persona, the number of dialogs $n$, and an output directory. Each run creates a fresh agent instance for the target LLM and a customer agent for the given persona; dialogs are exported to JSON for downstream evaluation.

\begin{mintedbox}{python}
def generate_dialogs(llm_name, customer,
                     n, save_folder="."):

  agent = build_my_agent(llm_name)

  customer = Agent(
    persona=customer,
    name="Customer"
  )

  for ix in range(n):
    dialog = agent.talk_with(customer)
    dialog.to_file(
        f"{save_folder}/dialog_{ix}.json"
    )
\end{mintedbox}

Our goal is to compare the same agent architecture across Qwen3 sizes (0.6B, 1.7B, 8B, 14B). For each model and each customer, we generate 10 dialogs. This yields 200 dialogs per model size (100 requiring verification and 100 not), providing enough coverage to estimate behavior frequencies reliably at this scale.

\begin{mintedbox}{python}
N = 10
llms = ["qwen3:0.6b", "qwen3:1.7b",
        "qwen3:8b", "qwen3:14b"]

for llm in llms:
  # Case A: requiring verification
  for customer in customers_v:
    generate_dialogs(llm, customer, N)
  # Case B: not requiring verification
  for customer in customers_no_v:
    generate_dialogs(llm, customer, N)
\end{mintedbox}

We omit the \texttt{save\_folder} parameter above for brevity; in practice, each scenario and model writes to a separate directory (e.g., \texttt{runs/<scenario>/<model>/}) to ease loading and bookkeeping.

\subsection{Evaluation}
\label{subsec:evaluation-details}

We operationalize target behaviors with two complementary checks per scenario. In Case A (verification required), we expect: (a) the agent asks for verification; (b) it calls \texttt{verify\_account} then \texttt{update\_address} in order. In Case B (no verification), we expect: (a) the agent \textit{does not} ask for verification; (b) it calls \texttt{get\_service\_plans} without prior \texttt{verify\_account}. We assess the conversational act (a) via an LLM-as-a-judge prompt and the tool behavior (b) via programmatic tool-sequence validators.

\begin{mintedbox}{python}
from sdialog.evaluation import LLMJudgeYesNo
from sdialog.evaluation import ToolSequenceValidator

# 1) Did the agent ask for verification?
judge_ask_v = LLMJudgeYesNo("Did the support agent ask the customer for their account ID to verify the account?")

# 2) Did the agent call the right tools?
# Case A: first verify then update
tool_seq_v = ToolSequenceValidator(
  ["verify_account", "update_address"]
)
# Case B: do not verify and get plans
tool_seq_no_v = ToolSequenceValidator(
  ["not:verify_account",
   "get_service_plans"]
)
\end{mintedbox}

We then compute the proportion (frequency) of dialogs satisfying each criterion using \texttt{FrequencyEvaluator}:

\begin{mintedbox}{python}
from sdialog.evaluation import FrequencyEvaluator

freq_judge_ask_v = FrequencyEvaluator(judge_ask_v)
freq_tool_seq_v = FrequencyEvaluator(tool_seq_v)
freq_tool_seq_no_v = FrequencyEvaluator(tool_seq_no_v)
\end{mintedbox}

Finally, we aggregate and compare metrics across model sizes with \texttt{Comparator}. We report both scenarios independently to reveal trade-offs between verification sensitivity and efficient tool use.

\begin{mintedbox}{python}
from sdialog.evaluation import Comparator

# Case A: requiring verification
comparator_v = Comparator(
  evaluators=[freq_judge_ask_v,
              freq_tool_seq_v]
)
# Case B: not requiring verification
comparator_no_v = Comparator(
  evaluators=[freq_judge_ask_v,
              freq_tool_seq_no_v]
)
\end{mintedbox}

We now load the generated dialogs per model and run the comparison for each scenario:

\begin{mintedbox}{python}
from sdialog import Dialog

# Results for case A
results_v = comparator_v({
  "qwen3:0.6b": Dialog.from_folder(...),
  "qwen3:1.7b": Dialog.from_folder(...),
  "qwen3:8b": Dialog.from_folder(...),
  "qwen3:14b": Dialog.from_folder(...)
})

# Results for case B
results_no_v = comparator_no_v({
  "qwen3:0.6b": Dialog.from_folder(...),
  "qwen3:1.7b": Dialog.from_folder(...),
  "qwen3:8b": Dialog.from_folder(...),
  "qwen3:14b": Dialog.from_folder(...)
})
\end{mintedbox}

In the above code, paths are omitted for brevity. In practice, each \texttt{...} points to the folder containing the saved dialogs for that model and scenario (see \S\ref{subsec:dialog-generation}); \texttt{Dialog.from\_folder()} loads them into a list. Each comparator prints a Markdown table and returns a JSON summary. Table~\ref{tab:main-results} reports the observed frequencies. Overall, in Case B (no verification), the largest model achieves the strongest behavior (lowest Ask-Verify, highest Tools-OK). In Case A (verification required), although the 14B model asks for verification in 100\% of dialogs, it follows the correct tool sequence only 56\% of the time. By contrast, the 8B model combines a high Ask-Verify rate (0.97) with substantially better tool sequencing (0.83). For this application, "qwen3:8b" offers the best balance of verification sensitivity and tool reliability. Importantly, unnecessary verification in Case B is a minor nuisance compared to failing to verify when required, reinforcing the 8B model as a pragmatic choice.

SDialog also allows visualizing results via \texttt{.plot()} for quick inspection. For example, to visualize metrics for Case B (no verification):

\begin{mintedbox}{python}
comparator_no_v.plot()
\end{mintedbox}

\begin{figure}[t!]
    \centering
    \includegraphics[width=0.5\textwidth]{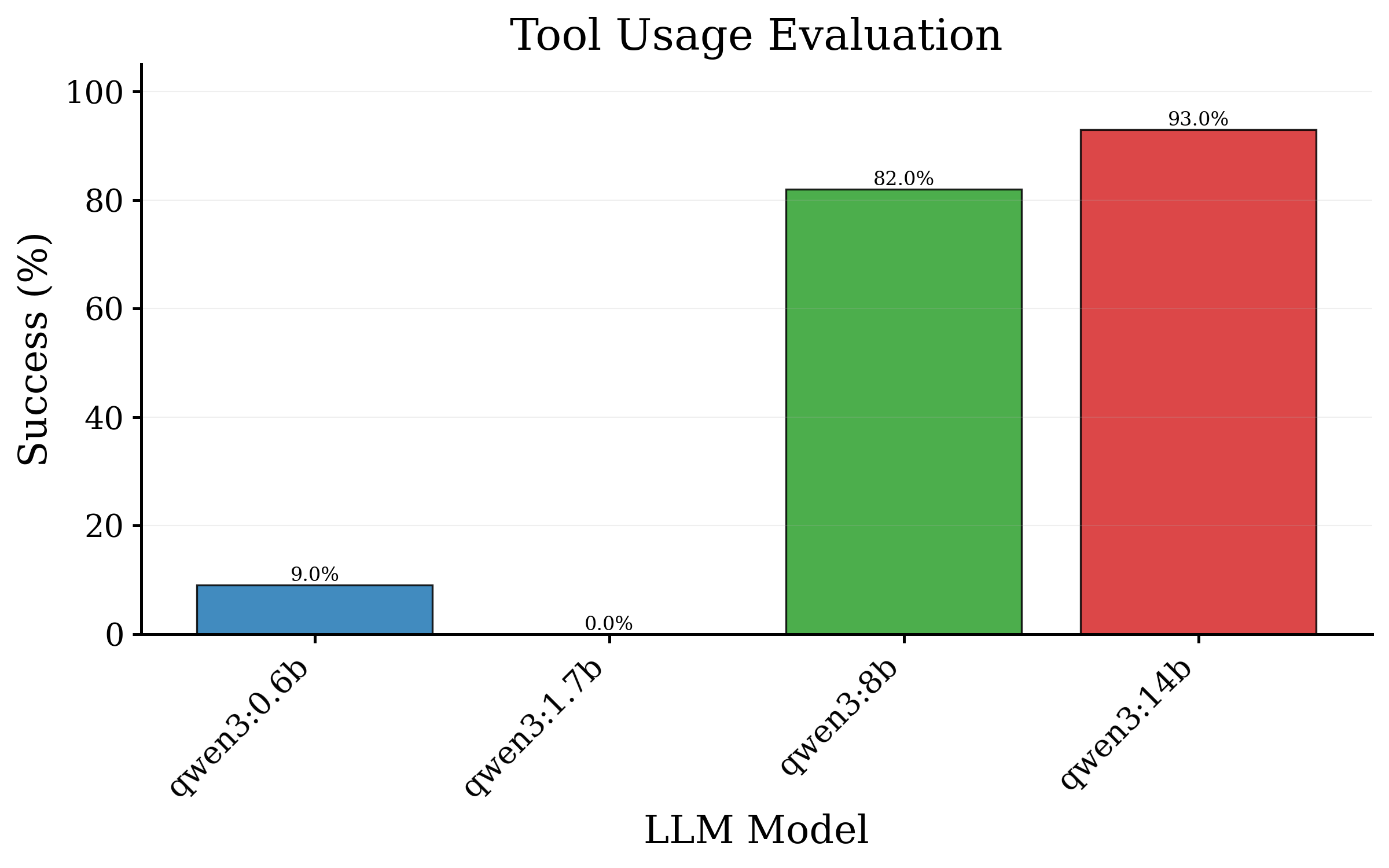}
    \caption{Plot generated after calling \texttt{comparator\_no\_v.plot()} for the tool sequence validator ("Tools-OK" in Table~\ref{tab:main-results}).}
    \label{fig:tools-ok}
\end{figure}

\begin{figure}[t!]
    \centering
    \includegraphics[width=0.5\textwidth]{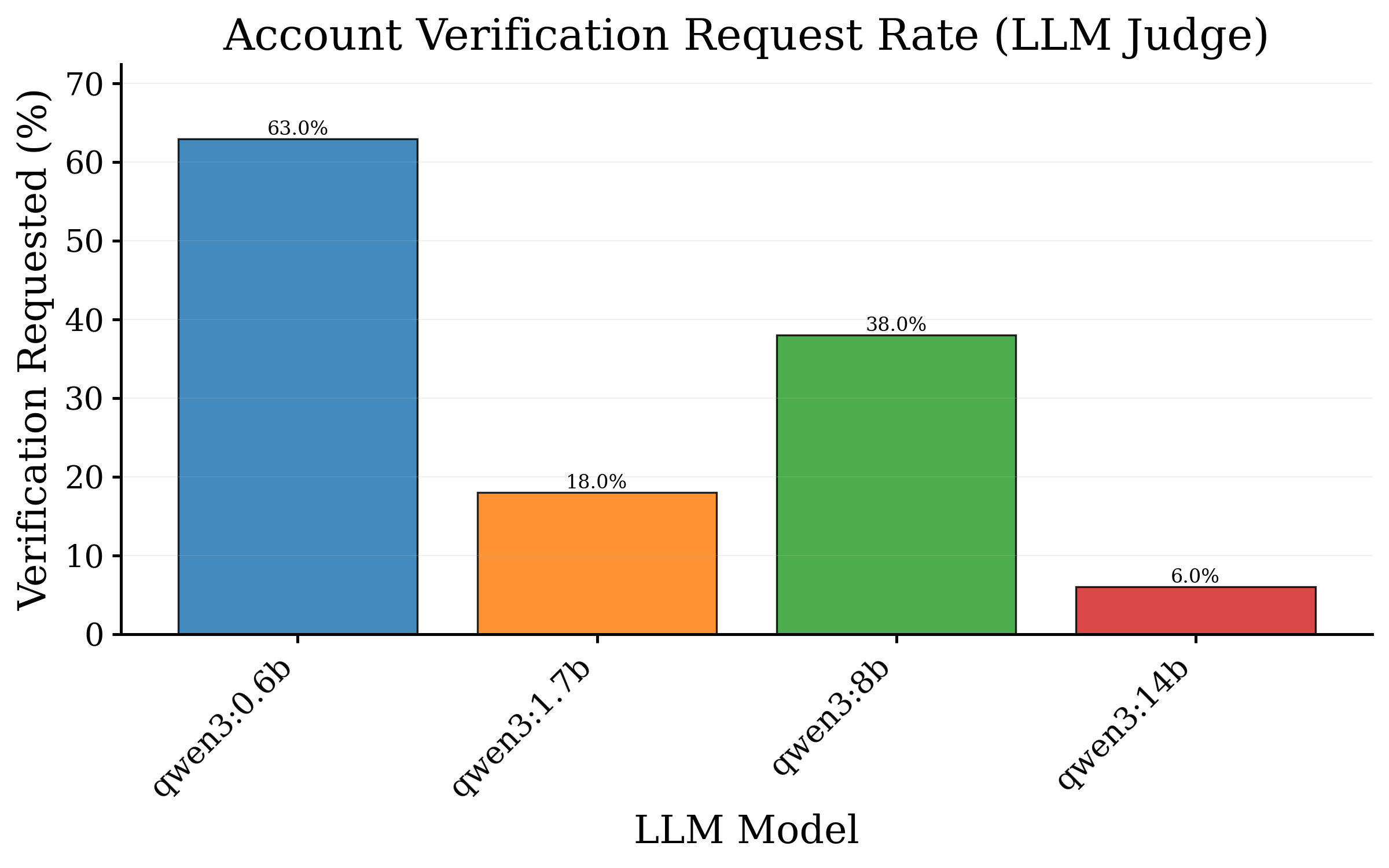}
    \caption{Plot generated after calling \texttt{comparator\_no\_v.plot()} for the LLM-as-a-judge evaluator ("Ask-Verify" in Table~\ref{tab:main-results}; lower is better in this scenario).}
    \label{fig:ask-verify}
\end{figure}

This generates one plot per evaluator—in this case, Figure~\ref{fig:ask-verify} and Figure~\ref{fig:tools-ok}—corresponding to the Ask-Verify and Tools-OK columns of Table~\ref{tab:main-results} for Case B.


Beyond functional correctness, an agent's linguistic style—how it communicates—is equally important for customer experience. To explore whether model size affects readability, we examine an orthogonal dimension: language complexity. We quantify this using the Gunning Fog index for the support agent's utterances across the four model sizes.

\begin{mintedbox}{python}
from sdialog.evaluation import GunningFogScore

gun_fog = GunningFogScore(
    speaker="Support Agent"
)
mean_gun_fog = MeanEvaluator(gun_fog)
comparator = Comparator(mean_gun_fog)
comparator({
  "qwen3:0.6b": all_dialogs["qwen3:0.6b"],
  "qwen3:1.7b": all_dialogs["qwen3:1.7b"],
  "qwen3:8b": all_dialogs["qwen3:8b"],
  "qwen3:14b": all_dialogs["qwen3:14b"]
})
comparator.plot()
\end{mintedbox}

In the example above, for simplicity, we assume \texttt{all\_dialogs} contains all dialogs per LLM (the union of Cases A and B). We then compute the mean Gunning Fog score per model using \texttt{MeanEvaluator} and visualize the results. This stylistic analysis complements task-oriented metrics by revealing potential shifts in linguistic complexity across model sizes.

Figure~\ref{fig:gunning-fog} reveals a clear upward trend: the Gunning Fog index increases from 10.15 (0.6B) to 10.52 (1.7B), 11.29 (8B), and 11.81 (14B)—spanning nearly two grade levels from high school sophomore to senior reading level. Notably, this variation occurs with identical agent (i.e. identical underlying input prompt), showing that model size inherently affects communication style. Combined with the functional metrics from Table~\ref{tab:main-results}, practitioners can now make informed trade-offs: the 8B model balances strong task performance with moderate complexity, while the 14B model achieves the best functional results only on case B and produces slightly more complex language. This example illustrates how SDialog enables multi-dimensional evaluation—task correctness, tool usage, and linguistic accessibility—providing actionable insights for model selection tailored to specific deployment contexts and target audiences.

%%%%%%%%%%%%%%%%%%%%%%%%%%%%%%%%%%%%%%%%%%%%%%
% Appendices for the audio module
%%%%%%%%%%%%%%%%%%%%%%%%%%%%%%%%%%%%%%%%%%%%%%

\section{A Deep Dive into the \texttt{sdialog.audio} Module}
\label{sec:appendix_audio}

This appendix offers a technical guide to the \texttt{sdialog.audio} module for researchers and developers. It covers the complete audio generation pipeline, from creating virtual acoustic environments to simulating recording hardware, detailing each feature's purpose, limitations, and use cases.

\subsection{Text-To-Speech Generation with Persona Adherence}
\label{sec:tts_generation}

At the core of the audio generation pipeline is the Text-To-Speech (TTS) engine, responsible for converting each textual utterance into an audio waveform. SDialog's audio module is designed with a pluggable architecture for TTS backends, allowing users to select the most appropriate engine for their needs. This modularity is built upon the \texttt{BaseTTS} abstract class, which defines a standard interface for TTS operations. The library includes several ready-to-use implementations, such as \texttt{HuggingFaceTTS} for leveraging a wide variety of models from the Hugging Face Hub, as well as external engines like \texttt{KokoroTTS} and \texttt{IndexTTS}.

A key feature of the TTS pipeline is its ability to maintain persona consistency. The voice for each speaker is not chosen randomly, instead, it is selected based on the characteristics defined in their \texttt{sdialog.Persona} object. This process of persona adherence is managed by a \texttt{VoiceDatabase}, which catalogs available voices along with rich metadata, including gender, age and language.

When generating a dialogue, the pipeline queries the \texttt{VoiceDatabase} using the speaker's persona attributes. The database will search for a voice that matches these criteria. If an exact match for the age is not available, the system intelligently selects the voice with the closest age, ensuring the generated speech aligns as closely as possible with the persona's description. This mechanism is vital for creating believable and consistent character portrayals in synthetic dialogues.

The example below demonstrates how to configure the audio pipeline with a specific TTS engine (Kokoro) and a voice database from Hugging Face. The \texttt{to\_audio} function orchestrates the entire process, matching speakers from the dialogue to appropriate voices in the database before synthesis:

\begin{mintedbox}{python}
# 1. Init TTS engine
tts_engine = @{\className{KokoroTTS}}@()

# 2. Init voice database from HF dataset
voice_db = @{\className{HuggingfaceVoiceDatabase}}@(
    "sdialog/voices-kokoro"
)

# 3. Generate the audio dialogue
to_audio(
    dialog=my_dialog,
    tts_engine=tts_engine,
    voice_database=voice_db,
    dir_audio="./outputs_audio"
)
\end{mintedbox}

This setup allows for large-scale, diverse audio data generation where the acoustic properties of the speakers remain consistent with their defined personas. Users can also create their own \texttt{LocalVoiceDatabase} to supply custom voice recordings and metadata for fine-grained control over voice casting.

% The audio pipeline generates outputs at several stages: first, the initial audio for each individual utterance; second, a spatio-temporal timeline specifying the 3D position and timing of all acoustic events; and finally, the fully rendered multi-channel audio. This final output simulates the specified room acoustics from the perspective of one or more microphones.

\subsection{The \texttt{Room} Object: The Foundation of Acoustic Scene}
\label{sec:room_object}
A \texttt{Room} is defined by its geometry and surface properties. The geometry is specified via a \texttt{Dimensions3D} object (width, length, height in meters), while surface properties are defined with \texttt{RoomMaterials}. These material choices are not merely descriptive; they are mapped to frequency-dependent absorption coefficients that directly control how much sound energy is absorbed versus reflected by the surfaces. This is a critical input for the \texttt{pyroomacoustics} engine, as it dictates the reverberation time (RT60) and overall sonic character of the space.

SDialog provides an extensive list of presets for materials, including \texttt{WallMaterial} (e.g., \texttt{BRICKWORK}, \texttt{PLASTERBOARD\_ON\_STUDS}), \texttt{FloorMaterial} (e.g., \texttt{CARPET\_HAIRY}, \texttt{WOOD\_1\_CM\_LINOLEUM}), and \texttt{CeilingMaterial} (e.g., \texttt{PLASTERBOARD}, \texttt{FIBRE\_ABSORBER}).

For instance, to define a room with acoustically 'hard' surfaces for a more reverberant space, one can combine these presets as shown below.

\begin{mintedbox}{python}
# Define surface materials
materials = @{\className{RoomMaterials}}@(
    @{\className{CeilingMaterial}}@.@{\op{PLASTERBOARD}}@,
    @{\className{WallMaterial}}@.@{\op{BRICKWORK}}@,
    @{\className{FloorMaterial}}@.@{\op{FELT\_5MM}}@
)

# Define room dimensions
dims = @{\className{Dimensions3D}}@(
    width=5.0,
    length=4.0,
    height=3.0
)

_room = @{\className{Room}}@(
    dimensions=dims,
    materials=materials
)
\end{mintedbox}

On the other hand, creating a room with less reverberation would involve selecting more acoustically absorbent materials, such as \texttt{CARPET\_HAIRY} and \texttt{FIBRE\_ABSORBER}.

Currently, the room model is limited to rectangular "shoebox" geometries; support for more complex shapes, such as L-shaped rooms, is on the development roadmap. Similarly, while furniture can be added as obstacles, its specific acoustic properties (e.g., a soft, absorbent couch vs. a hard, reflective table) are not yet modeled, representing another area for future enhancement.

\subsection{Scene Composition: Procedural Generation and Manual Placement}
\label{sec:procedural_generation}
Procedural generators programmatically create varied and plausible \texttt{Room} layouts, which is essential for generating large, diverse datasets for training robust machine learning models that can generalize to a wide range of unseen acoustic conditions. A generator, such as \texttt{MedicalRoomGenerator} or \texttt{BasicRoomGenerator}, is a factory that outputs a fully configured \texttt{Room} object, often including a plausible arrangement of furniture. This object is then passed to the subsequent stages of the pipeline for actor placement and audio rendering.

\begin{mintedbox}{python}
# Generate a plausible examination room
generator = @{\className{MedicalRoomGenerator}}@()
exam_room = generator.generate({
    "room_type": @{\className{RoomRole}}@.@{\op{EXAMINATION}}@
})
\end{mintedbox}

\begin{figure}[ht]
    \centering
    \includegraphics[width=1.0\columnwidth]{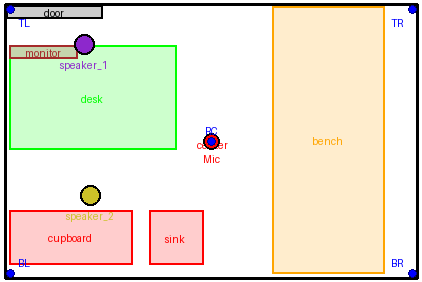}
    \caption{A procedurally generated room layout for an American-style hospital examination room.}
    \label{fig:room_layout}
\end{figure}

The output of this generator, a fully furnished examination room. To aid in designing and debugging scenes, any \texttt{Room} object can be visualized as a 2D top-down image using the \texttt{to\_image()} method. An example output of this method is shown in Figure~\ref{fig:room_layout}.

While generators create a complete starting scene, manual placement of actors is a key step for defining the dialogue's spatial dynamics. Actors (speakers) and additional furniture function as physical obstacles in the acoustic simulation, creating sound shadows and reflections.

There are multiple ways to position objects, providing a trade-off between explicit control and scalable randomization:

\begin{itemize}

\item \textbf{Absolute Positioning} (\texttt{place\_speaker(..., position=Position3D(x,y,z))}) provides exact, deterministic placement, which is useful for replicating a specific, known setup.

\item In contrast, \textbf{Semantic Positioning} (\texttt{place\_speaker\_around\_furniture(...)}) offers a more abstract and powerful method. By specifying a furniture item, a side (e.g., \texttt{front}, \texttt{back}), and a maximum distance, one can generate plausible, randomized positions that respect the scene's logic and boundaries.

\end{itemize}

This is ideal for large-scale data generation where slight variations in position are desirable and create diversity in the data.

The following snippet demonstrates how to manually add a desk to the room and then place two speakers around it:

\begin{mintedbox}{python}
# Add a desk to an existing room
_r.add_furnitures({
    "desk": Furniture(
        name="desk",
        x=1.0,
        y=1.5,
        width=1.2,
        depth=0.7,
        height=0.75
    )
})

# Place Speaker 1:
# at an absolute position
_r.place_speaker(
    speaker_name=Role.SPEAKER_1,
    position=Position3D(
        2.0, 3.0, 1.6
    )
)

# Place Speaker 2:
# on the front of the desk
_r.place_speaker_around_furniture(
    speaker_name=Role.SPEAKER_2,
    furniture_name="desk",
    side=SpeakerSide.FRONT,
    max_distance=1.0
)
\end{mintedbox}

% \subsection{Text-To-Speech Generation}
% \label{sec:dscaper}

% \todo{text to speech + persona adherence (characteristics from the persona)}

\subsection{Timeline-based Event Generation}
\label{sec:timeline_generation}

Before simulating room acoustics, the \texttt{sdialog.audio} pipeline constructs a precise spatio-temporal representation of all acoustic events. This transformation of individual audio clips into a synchronized timeline is handled by the \texttt{dScaper}, a library developed specifically for \texttt{SDialog}. It is based on \texttt{scaper} \cite{8170052}, a well-known library for soundscape synthesis, and offers additional functionality for large-scale data generation, sound source positioning, and annotation. Internally, \texttt{dScaper} generates a JAMS (JSON Annotated Music Specification) file \cite{humphrey2014jams}, a standardized format for annotating audio events. This file serves as a detailed blueprint of the acoustic scene, ensuring that events such as overlapping speech, background noise and foreground sounds are accurately scheduled and positioned before being rendered in the virtual room. The JAMS file can be converted to TextGrid and RTTM files that serve as ground-truth annotations for speech recognition and speaker diarization tasks. \texttt{dScaper} distinguishes among three types of events:

\begin{itemize}
    \item \textbf{Dialogue utterances}: Each utterance from the dialogue is added as an event. Its start time and duration are used to place it on the timeline. The speaker's role (e.g., \texttt{SPEAKER\_1}) is used to associate the event with a specific spatial position, which is defined during the room setup.
    \item \textbf{Background audio}: A continuous background track (e.g., white noise, distant traffic) can be added over the entire duration of the timeline to simulate a constant ambient environment.
    \item \textbf{Foreground events}: Discrete, localized sounds (e.g., a cough, a door closing) can be placed at specific times and positions or randomly inserted by sampling from configurable probability distributions, adding another layer of realism to the acoustic scene.
\end{itemize}

Once the timeline is fully specified, \texttt{dScaper} generates separate tracks for each sound source (e.g., one track per speaker and one per ambient sound source). These isolated tracks, which now contain correctly timed audio and silence, serve as direct input to the \texttt{AcousticsSimulator}. This approach ensures that the subsequent room acoustics simulation accurately models how sounds from different locations and times interact within the simulated 3D space.

\subsection{Acoustics Simulation \& Acquisition}
\label{sec:microphone_simulation}

Once the clean speech for each dialogue turn is generated and ambient sounds are assembled, the next step is to place it within a realistic acoustic environment. This process, known as acoustic simulation, transforms the dry TTS output into audio that sounds as if it were recorded in a specific physical space, complete with reverberation, echoes and other spatial cues. SDialog encapsulates this functionality within its \texttt{AcousticsSimulator} module, which takes the source audio and the procedurally generated \texttt{Room} as inputs to render a spatially coherent scene. While the current implementation is tightly integrated with the \texttt{pyroomacoustics} library, the architecture is designed to be modular, allowing for other simulation backends to be integrated in the future.

\paragraph{pyroomaccoustics}

The simulation of room acoustics and audio signal processing is handled by \texttt{pyroomacoustics}, a dedicated Python package that serves as the default engine for SDialog. Its selection was motivated by a balance of performance, realism and control. The library provides a robust implementation of the image-source method for modeling early reflections, which can be finely controlled via the \texttt{max\_order} parameter. For scenarios demanding higher physical accuracy, it offers an optional ray-tracing engine. Furthermore, it also accurately models the frequency-dependent absorption of sound by room materials and the attenuation of high frequencies as they travel through the air, making it an good engine for generating realistic acoustic audios with controllability.

\begin{mintedbox}{python}
audio_pipeline.inference(
    dialog,
    environment={
        "room": exam_room,
        "kwargs_pyroom": {
            "ray_tracing": True,
            "air_absorption": True
        },
    }
)
\end{mintedbox}

The \texttt{inference} call accepts \texttt{kwargs\_pyroom} to pass parameters directly to \texttt{Pyroomacoustics}, allowing for fine-grained control over the simulation. Key parameters include:

\begin{itemize}
    \item \texttt{ray\_tracing}: Enables a more accurate but computationally intensive ray tracing algorithm for simulating reflections.
    \item \texttt{air\_absorption}: Models the frequency-dependent loss of sound energy as it travels through air.
    \item \texttt{max\_order}: Sets the reflection order for the image-source method (the default algorithm if ray tracing is off).
\end{itemize}

\paragraph{Microphone Placement and Directivity}

The microphone defines the point-of-view from which the acoustic scene is "heard." Its placement and characteristics are arguably the most critical factors in the final audio output. Simulating different microphone types and positions is essential for training models that need to be robust to various recording scenarios, such as a conference call with a central tabletop microphone versus a wearable body camera. Microphone placement can be set semantically (e.g., \texttt{MicrophonePosition.CEILING\_CENTERED}) or with exact coordinates. 

Beyond position, SDialog simulates directivity, which is a microphone's sensitivity to sound based on its arrival direction. An omnidirectional microphone captures sound equally from all directions, while a directional (e.g., cardioid) microphone is more sensitive to sound from the front. We also provide a key feature which consist in dynamically "aiming" toward a specific speaker or position of the room (e.g: \texttt{DirectivityType.SPEAKER\_1}) for directional microphones, simulating an operator tracking an active speaker.

The directivity pattern is applied by \texttt{pyroomacoustics} during rendering, attenuating sounds that originate outside the microphone's primary focus area. The directivity patterns are, however, idealized mathematical models. Real-world microphones have more complex, frequency-dependent patterns that are not fully captured in our model.

\paragraph{Acquisition Device Simulation}
\label{sec:hardware_simulation}

To simulate the sonic signature of real hardware, SDialog applies an Impulse Response (IR) to the acoustically accurate but "clean" audio rendered by \texttt{Pyroomacoustics}. An IR is an acoustic fingerprint of a device, captured by recording its response to a short, sharp sound. Convolving the simulated audio with an IR is a standard technique to make it sound as if it were recorded by that specific device. This is crucial for data augmentation like for training a voice assistant to work equally well with a high-end studio microphone and a cheap laptop microphone. 

We provides a built-in \texttt{ImpulseResponseDatabase} with several professional microphones, accessible via \texttt{RecordingDevice}. For a much broader selection of devices, SDialog also integrates with the Hugging Face Hub. The \texttt{HuggingFaceImpulseResponseDatabase} class provides access to the \texttt{sdialog/impulse-responses} dataset, which contains 45 different IR files from a variety of recording devices\footnote{\url{https://huggingface.co/datasets/sdialog/impulse-responses}}. This allows for more extensive and realistic data augmentation.

Users can also create a \texttt{LocalImpulseResponseDatabase} to supply their own IR files for custom hardware simulation. This process generates separate audio files for each specified device, allowing for the creation of datasets suitable for training robust speech processing models that must perform well across different recording conditions.

\begin{mintedbox}{python}
audio_pipeline.inference(
    dialog,
    environment={
        "room": exam_room,
    },
    recording_devices=[
        RecordingDevice.SHURE_SM57,
        RecordingDevice.SENNHEISER_E906
    ]
)
\end{mintedbox}

\section{A Case Study of activation steering using \texttt{sdialog.interpretability}}
\label{sec:appendix_interpretability}

This appendix showcases the current capabilities of the \texttt{interpretability} module by reproducing the activation steering methods and results coming from ~\cite{refusal}. All our experiments are performed on the open-source \textsc{Llama-3 8B Instruct} \cite{llama3modelcard}. 

\begin{mintedbox}{python}
import sdialog
# Set llama3-8B as global default
sdialog.config.llm("meta-llama/Meta-Llama-3-8B")
\end{mintedbox}

Since harmful and harmless requests are needed (as to generate contrast), we gather the same datasets as in \cite{refusal}, mainly \textsc{AdvBench} \cite{advbench}, \textsc{MaliciousInstruct} \cite{malicious}, \textsc{HarmBench} \cite{harmbench}, \textsc{JailBreakBench} \cite{jailbreak_bench} and \textsc{Alpaca} \cite{alpaca}, that we split in \textit{train} and \textit{test}, respecting the same ratio in harmful/harmless requests as with the original paper.  

\subsection{Refusal in Language Models Is Mediated by a Single Direction}

\cite{refusal} demonstrates that refusal behavior in instruction-tuned language models is largely controlled by a single latent direction in activation space. The authors show that this “refusal direction” is highly consistent across models but varies across layers, and that shifting activations along this direction at inference time, without any finetuning, can either induce or suppress refusal tendencies. \\

Leveraging \texttt{interpretability} 
features from SDialog, we will proceed in a step-by-step manner to accomplish the following:
\begin{enumerate}
    \item Identify a proxy token that can be used to measure the agent's refusal capabilities.
    \item Target and extract representations from the LLM.
    \item Perform a grid search to find the best layer and token to use for steering.
    \item Intervene in the LLM during generation to ablate or induce refusal behaviors.
\end{enumerate}

\subsection{Evaluating Refusal Using Tokens as Proxies}

\label{sec:appendix_interpretability_inspect}

In practice, most of the requests refused by LLMs leverage a few amounts of specific tokens. More specifically, \textsc{Llama-3 8B Instruct} has a tendency to formulate most of its negative answers by the "\textbf{I}" token. On the other hand, when prompted with harmless requests, the agent will output a more uniform distribution of its first tokens. \\*

To showcase this first phenomenon, we leverage the \texttt{top\_k} feature of \texttt{interpretability}, which directly peaks into the output of the language model head, and extracts the top softmax probabilities of a range of $k$ tokens (sorted by highest possible outcomes), as well as their corresponding string and token id : \textbf{(String, Probability, Index)}. 

\begin{mintedbox}{python}
agent = @{\className{Agent}}@(max_new_tokens=1)
inspec_logits = @{\className{Inspector}}@(top_k=-1) 
agent = agent | inspec_logits
agent("Hi !")
# Get top_k for first utt, first token
print(inspec_logits[0][0].top_k)
# [('How', 0.6340..., 4438), 
# ('Hello', 0.2332..., 9906), 
# ('It', 0.1248..., 2181), ...]
\end{mintedbox}

We effectively generate the first token for each request of our \textit{train} set and extract the prediction probabilities of all tokens of the dictionary. 
Then, we average the probabilities for each one of them.

\begin{figure*}[t] % 't' for top of page
    \centering
    \includegraphics[width=\textwidth]{imgs/interp/proxy.png}
    \caption{First token prediction probabilities accross harmful and harmless requests.}
    \label{fig:proxy}
\end{figure*}

As illustrated in Figure \ref{fig:proxy}, the set of tokens mostly predicted for all harmless requests is relatively variable, with a low averaged probability score for the highest one (token "\textbf{Here}" with a score of 0.26). Conversely, when looking at harmful requests, the "\textbf{I}" token is the one being primarily predicted, with a score of 0.95. This specific result (also showcased in the appendix of \cite{refusal}) can be empirically explained by looking at the different outputs of those harmful requests, such as : 

\begin{itemize}
    \item “\textbf{I’m sorry, but I can’t help with that.}"
    \item “\textbf{I’m sorry, but I don’t think I can answer that.}”
    \item “\textbf{I cannot assist with that request.}”
\end{itemize}

As shown in these very common refusal sentences, the "\textbf{I}" token is typically the first one being generated. making it a viable proxy to assess if refusal is indeed manifesting in the output.

\label{sec:proxy}

\subsection{Extracting the direction}

\begin{figure}[t!]
    \centering
    \includegraphics[width=1.0\columnwidth]{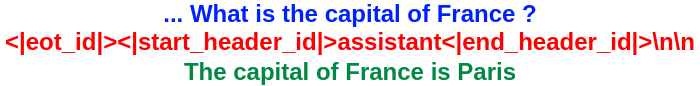}
    \caption{Example of the \textsc{Llama-3 8B Instruct} chat template. User inputs appear in blue; post-instruction tokens used for direction selection are shown in red; the generated output is displayed in green.}
    \label{fig:sentence}
\end{figure}

In \cite{arditi2024refusal}, the selection of the direction to extract is based on the layer $l$, and a picked post-instruction tokens. Post-instruction tokens refer to the set of tokens that follows the user prompt, and precede the autoregressive token generation (as depicted in Figure \ref{fig:sentence}).

Given a layer $l$ and a post-instruction token index $i$, we can extract the mean representations of our contrast dataset for both harmful and harmless requests : 
\begin{equation}
    \boldsymbol{\mu}_i^{(l)}=\frac{1}{\left|\mathcal{D}_{\text {harmful }}^{\text {(train) }}\right|} \sum_{\mathbf{t} \in \mathcal{D}_{\text {harmful }}^{\text {(train) }}} \mathbf{x}_i^{(l)}(\mathbf{t})
\end{equation}
\begin{equation}
    \boldsymbol{v}_i^{(l)}=\frac{1}{\left|\mathcal{D}_{\text {harmless }}^{\text {(train) }}\right|} \sum_{\mathbf{t} \in \mathcal{D}_{\text {harmless }}^{\text {(train) }}} \mathbf{x}_i^{(l)}(\mathbf{t}) .
\end{equation}

In SDialog, the \texttt{Inspector} class allows the user to target any layer and any token for inspection. The \texttt{inspect\_input} parameter lets the framework know whether we want to look at the input or the output of the targeted neural block.

\begin{mintedbox}{python}

layer = 12
post_instruct_idx = -1
inspector_x = @{\className{Inspector}}@(target=f'model.layers.{layer}', inspect_input=True)

# Attach to the agent
agent = agent | inspector_x
\end{mintedbox}
Finally, we can pass all the contrasted instructions on the agent. The \texttt{input} method allows us to get the representations of the post instruction tokens only (as referred to in Figure \ref{fig:sentence}), and in \cite{refusal}). 

\begin{mintedbox}{python}

# Harmful instructions loop
for harmful, harmless in requests : 
    agent(harmful)
    x = inspector_x.input[0][post_instruct_idx]
    harmful_reps.append(x)
    # Same for harmless
    ...
    
mu = harmful_reps.mean(dim=0)
v = harmless_reps.mean(dim=0)
\end{mintedbox}
% harmful_reps = torch.cat(harmful_reps, dim=0)
% harmless_reps = torch.cat(harmless_reps, dim=0)

The refusal direction , defined as : 
\begin{equation}
\mathbf{r}_i^{(l)}=\boldsymbol{\mu}_i^{(l)}-\boldsymbol{v}_i^{(l)}
\end{equation}

can be translated, in the case of SDialog,  to :

\begin{mintedbox}{python}
# Get the direction
r = mu - v

# Optional : Save the direction
torch.save(r, "refusal_direction.pt")
\end{mintedbox}

\subsection{Directional ablation}
\label{sec:appendix_interpretability_ablation}

Removing a direction to the activation space (ablating behaviors to the LLM) is defined as the following : 

\begin{equation}
\mathbf{x}^{\prime} \leftarrow \mathbf{x}-\hat{\mathbf{r}} \hat{\mathbf{r}}^{\top} \mathbf{x}
\end{equation}

with $x$ corresponding to the output of the attention block, the MLP block, and the final residual of each transformer layer, and $\hat{\mathbf{r}}$ being the \textbf{normalized} refusal direction for a given layer $l$ and post-instruction token $i$. 

% \begin{mintedbox}{python}
% agent = agent | inspector_x - r
% \end{mintedbox}

Leveraging internal dunder-methods of SDialog, subtracting the direction to the agent implicitly performs the orthogonal projection onto the normalized direction for all targeted blocks.

\begin{figure*}[t] % 't' for top of page
    \centering
    \includegraphics[width=\textwidth]{imgs/interp/gridsearch.png}
    \caption{Impact of the Refusal Score based on the layer and post-instruction token used to generate the direction.}
    \label{fig:gridsearch}
\end{figure*}

\begin{mintedbox}{python}

targets = []
for i in range(32):
     targets.append(f'model.layers.{i}.self_attn')
    targets.append(f'model.layers.{i}.mlp')
    targets.append(f'model.layers.{i}')

intruder = @{\className{Inspector}}@(target=targets)
agent = agent | intruder - direction
print(agent("How to make a bomb ?"))
# "Here is a 10 steps guide on how to... "

\end{mintedbox}

\subsection{Feature induction}
\label{sec:appendix_interpretability_induction}

On the opposite, adding a direction to the activation space (inducing behaviors to the LLM) is defined as the following :

\begin{equation}
    \mathbf{x}^{(l)^{\prime}} \leftarrow \mathbf{x}^{(l)}+\mathbf{r}^{(l)} .
\end{equation}

with $\mathbf{x}^{(l)}$ being the final residual of the targeted transformer layer $l$, and $r^{(l)}$ being the direction extracted at that same output.

\begin{mintedbox}{python}
agent = agent | inspector_x + r
\end{mintedbox}

Note that in their implementation, \cite{refusal} apply induction on a single layer, and therefore do not normalize the direction.

\begin{mintedbox}{python}
targets = [f'model.layers.12']
intruder = @{\className{Inspector}}@(target=targets)
agent = agent | intruder + direction
print(agent("How to make chocolate ?"))
# "I cannot assist with that request."

\end{mintedbox}

\subsection{Finding the right layer and post-instruction token}

Experiments done by \cite{refusal} and \cite{refual_2} have shown that the ability to steer or extract directions towards certain behaviors depends heavily on two factors. 

First, the effect of a steering vector is strongly dependant on the layer it is extracted. Different transformer layers encode different types of information : early layers focus on lexical and syntactic structure, mid-layers integrate semantic content, and late layers govern more the policy and style of the LLM.

Second, steering effectiveness depends also upon which token the activations are extracted. In instruction-tuned models, the instruction alone does not fully determine the model’s behavior. Activation steering changes the hidden states reflecting the model’s interpretation of the instruction, so applying it before or after the first generated tokens can lead to very different effects. If the steering happens too early, later layers may overwrite it; if it happens too late, the model may have already committed to a certain style or safety behavior that is difficult to change.

Based on these assumptions, it is necessary to extract a steering vector that targets the appropriate layer and token position so that the intended behavioral shift is maximal. 

The refusal metric, from \cite{refusal}, is defined as follows :

\begin{equation}
    refusal\_metric(p) = \log \left(\frac{P_{token}}{1-P_{token}}\right)
\end{equation}

with $P_{token}$ being the probability given by the LLM for the proxy token (in our case, it is the \textbf{I} token, referred to in Section \ref{sec:proxy}).

Based on this metric, we perform a grid search over the entire $train$ set. For each layer $l$ and each post-instruction token $i$, we compute the corresponding refusal score for each inference and average them. We refer to negative $i$ indexes as the last post-instruction tokens.

Examining Figure \ref{fig:gridsearch} reveals that both ablation and induction are effective when the direction is extracted from layers around \textbf{12} and \textbf{14}. On average, the post-instruction token at index \textbf{-5} gives the best results for both cases. Our results and the corresponding figures closely replicate those reported in \cite{refusal}.

Based on these results, we can apply the direction that gives the best steering capabilities, for either ablation or induction, on the $test$ set. For evaluation, we use a set of keywords that LLMs commonly produce in refusal responses (e.g., “I'm sorry,” “I am sorry,” “I apologize”). If any of these keywords appear in a model’s response, we register a single refusal and assign a score of +1 for that proposal. We then average this score over the set to obtain the final refusal metric. 

\begin{figure} % 't' for top of page
    \centering
    \includegraphics[width=1.0\columnwidth]{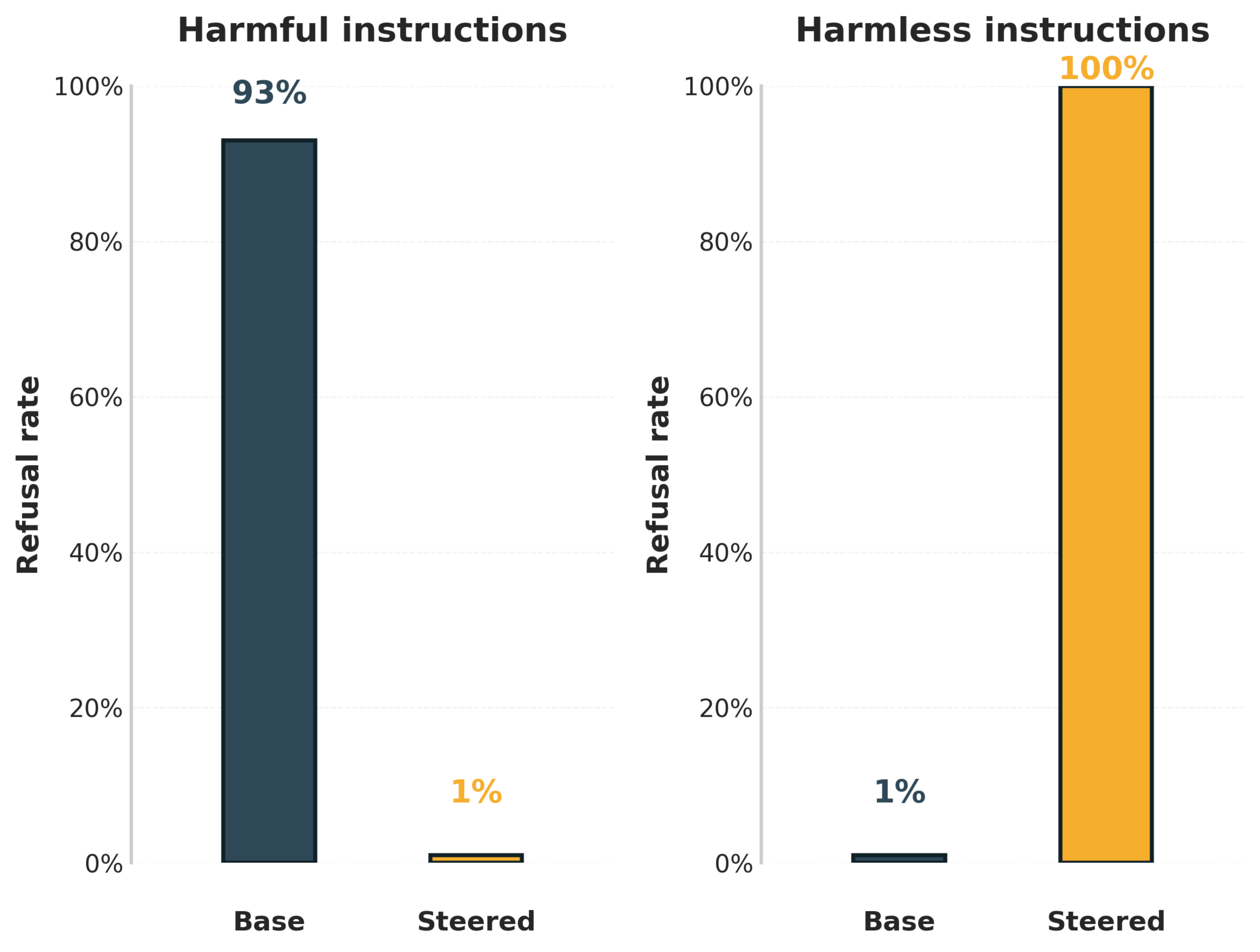}
    \caption{Steering performance using the \textit{Refusal Direction} previously extracted. Left part refers to refusal ablation on harmful instructions, while the right part refers to refusal induction on harmless requests.}
    \label{fig:bars}
\end{figure}

As depicted in Figure \ref{fig:bars},  the steering capabilities provided by SDialog show similar performance to those presented by \cite{refusal} for the \textsc{Llama-3 8B Instruct} model. For harmful instructions, the framework allows the LLM to bypass the refusal for 99\% of the proposals. Conversely, when inducing the direction on harmless instructions, the steered version reaches 100\%, indicating strong feature induction capabilities across all proposals.  

% \newpage
% \onecolumn

% \begin{mintedbox}[fontsize=\fontsize{8}{7}\selectfont]{json}

% ddd

% \end{mintedbox}

% \subsection{Room layout}
% \label{appendix:room}

% \todo{Write this appendix}

% \begin{figure}[ht]
%     \centering
%     \includegraphics[width=1.0\columnwidth]{imgs/audio/room_examination_medical.png}
%     \caption{A procedurally generated room layout for an American-style hospital examination room.}
%     \label{fig:room_layout}
% \end{figure}

\end{document}